\documentclass[review,12pt,3p]{elsarticle}
\usepackage{titlesec}

\titleformat{\paragraph}
{\normalfont\normalsize}{\theparagraph}{1em}{}
\titlespacing*{\paragraph}
{0pt}{3.25ex plus 1ex minus .2ex}{1.5ex plus .2ex}

\usepackage{setspace}
\usepackage{algorithm}
\usepackage{algorithmic}
\usepackage{boxedminipage} 
\usepackage{graphicx}
\usepackage{verbatim} 
\usepackage{amsmath}
\usepackage{times} 
\usepackage{gensymb}
\usepackage{dsfont}
\usepackage{multirow}
\usepackage{amsfonts}
\usepackage[T1]{fontenc}
\usepackage{url}
\usepackage{amssymb}
\usepackage[tight,footnotesize]{subfigure}
\usepackage[caption=false,font=footnotesize]{subfig}
\usepackage{epsfig}

\begin{document}

\begin{frontmatter}

\title{META-DES: A Dynamic Ensemble Selection Framework using Meta-Learning}		

\author[ets]{Rafael M. O. Cruz\corref{corr1}}
\ead{rmoc@cin.ufpe.br}

\author[ets]{Robert Sabourin}
\ead{robert.sabourin@etsmtl.ca}

\author[ufpe]{George D. C. Cavalcanti}
\ead{gdcc@cin.ufpe.br}

\author[ufpe]{Tsang Ing Ren}
\ead{tir@cin.ufpe.br}

\address[ets]{LIVIA, \'{E}cole de Technologie Sup\'{e}rieure, University of Quebec, Montreal, Que., Canada - www.livia.etsmtl.ca
			 }
\address[ufpe]{Centro de Inform\'{a}tica, Universidade Federal de Pernambuco, Recife, PE, Brazil - www.cin.ufpe.br/$\sim$viisar\\
       }
			
\cortext[corr1]{Corresponding author. Email Address: rafaelmenelau@gmail.com.}

\begin{abstract}		

Dynamic ensemble selection systems work by estimating the level of competence of each classifier from a pool of classifiers. Only the most competent ones are selected to classify a given test sample. This is achieved by defining a criterion to measure the level of competence of a base classifier, such as, its accuracy in local regions of the feature space around the query instance. However, using only one criterion about the behavior of a base classifier is not sufficient to accurately estimate its level of competence. In this paper, we present a novel dynamic ensemble selection framework using meta-learning. We propose five distinct sets of meta-features, each one corresponding to a different criterion to measure the level of competence of a classifier for the classification of input samples. The meta-features are extracted from the training data and used to train a meta-classifier to predict whether or not a base classifier is competent enough to classify an input instance. During the generalization phase, the meta-features are extracted from the query instance and passed down as input to the meta-classifier. The meta-classifier estimates, whether a base classifier is competent enough to be added to the ensemble. Experiments are conducted over several small sample size classification problems, i.e., problems with a high degree of uncertainty due to the lack of training data. Experimental results show the proposed meta-learning framework greatly improves classification accuracy when compared against current state-of-the-art dynamic ensemble selection techniques.

\end{abstract} 

\begin{keyword}

Ensemble of Classifiers, Dynamic Ensemble Selection, Meta-Learning, Classifier competence

\end{keyword}

\end{frontmatter}

\section{Introduction}

Multiple Classifier Systems (MCS) aim to combine classifiers to increase the recognition accuracy in pattern recognition systems~\cite{kittler,kuncheva}. MCS are composed of three phases~\cite{Alceu2014}: (1) Generation, (2) Selection and (3) Integration. In the first phase, a pool of classifiers is generated. In the second phase, a single classifier or a subset having the best classifiers of the pool is(are) selected. We refer to the subset of classifiers as Ensemble of Classifiers (EoC). The last phase is the integration, and the predictions of the selected classifiers are combined to obtain the final decision~\cite{kittler}.

For the second phase, there are two types of selection approaches: static and dynamic. In static approaches, the selection is performed during the training stage of the system. Then, the selected classifier or EoC is used for the classification of all unseen test samples. In contrast, dynamic ensemble selection approaches (DES)~\cite{knora,docs,paulo2,logid,Woloszynski,dceid,ijcnn2011,mcb,lca,selectionfusion} select a different classifier or a different EoC for each new test sample. DES techniques rely on the assumption that each base classifier is an expert in a different local region of the feature space~\cite{zhu}. So, given a new test sample, DES techniques aim to select the most competent classifiers for the local region in the feature space where the test sample is located. Only the classifiers that attain a certain competence level, according to a selection criterion, are selected. Recent work in the dynamic selection literature demonstrates that dynamic selection techniques is an effective tool for classification problems that are ill-defined, i.e., for problems where the size of the training data is small and there are not enough data available to model the classifiers~\cite{paulo2,logid}.

The key issue in DES is to define a criterion to measure the level of competence of a base classifier. Most DES techniques~\cite{knora,lca,mcb,ijcnn2011,singh,Smits_2002,clussel,SoaresSCS06} use estimates of the classifiers' local accuracy in small regions of the feature space surrounding the query instance as a search criterion to perform the ensemble selection. However, in our previous work~\cite{ijcnn2011}, we demonstrated that the use of local accuracy estimates alone is insufficient to achieve results close to the Oracle performance. The Oracle is an abstract model defined in~\cite{Kuncheva:2002} which always selects the classifier that predicted the correct label, for the given query sample, if such classifier exists. In other words, it represents the ideal classifier selection scheme. In addition, as reported by Ko et al.~\cite{knora}, addressing the behavior of the Oracle is much more complex than applying a simple neighborhood approach. 

On the other hand, DES techniques based on other criteria, such as the degree of consensus of the ensemble classifiers~\cite{docs,paulo2}, encounter some problems when the search cannot find a consensus among the ensembles. In addition, they neglect the local performance of the base classifiers. As stated by the ``No Free Lunch'' theorem~\cite{freelunch}, no algorithm is better than any other over all possible classes of problems. Using a single criterion to measure the level of competence of a base classifier is very error-prone. Thus, we believe that multiple criteria to measure the competence of a base classifier should be taken into account in order to achieve a more robust dynamic ensemble selection technique. 

In this paper, we propose a novel dynamic ensemble selection framework using meta-learning. From the meta-learning perspective, the dynamic ensemble selection problem is considered as another classification problem, called meta-problem. The meta-features of the meta-problem are the different criteria used to measure the level of competence of the base classifier. We propose five sets of meta-features in this paper. Each set captures a different property about the behavior of the base classifier, and can be seen as a different dynamic selection criterion such as, the classification performance in a local region of the feature space and the classifier confidence for the classification of the input sample. Using five distinct sets of meta-features, even though one criterion might fail due to problems in the local regions of the feature space~\cite{ijcnn2011} or due to low confidence results~\cite{dosSantos}, the system can still achieve a good performance as other meta-features are also considered by the selection scheme. Furthermore, in a recent analysis~\cite{Cruz2014ANNPR} we compared the criteria used to measure the competence of base classifiers embedded in different DES techniques. The result demonstrates that, given the same query sample, distinct DES criteria select a different base classifier as the most competent one. Thus, they are not fully correlated. Hence, we believe that a more robust dynamic ensemble selection technique is achieved using five sets of meta-features rather than only one. 

The meta-features are used as input to a meta-classifier that decides whether or not a base classifier is competent enough for the classification of an input sample based on the meta-features. The use of meta-learning has recently been proposed in~\cite{Krijthe2012} as an alternative for performing classifier selection in static scenarios. We believe that we can carry this further, and extend the use of meta-learning to dynamically estimate the level of competence of a base classifier.

The proposed framework is divided into three phases: overproduction, meta-training and generalization. In the overproduction stage, a pool of classifiers is generated using the training data. In the meta-training stage, the five sets of meta-features are extracted from the training data, and are used to train the meta-classifier that works as the classifier selector. During the generalization phase, the meta-features are extracted from the query instance and passed down as inputs to the meta-classifier. The meta-classifier estimates whether a base classifier is competent enough to classify the given test instance. Thus, the proposed system differs from the current state-of-the-art dynamic selection techniques not only because it uses multiple criteria to perform the classifier selection, but also because the classifier selection rule is learned by the meta-classifier using the training data.

The generalization performance of the system is evaluated over 30 classification problems. We compare the proposed framework against eight state-of-the-art dynamic selection techniques as well as static combination methods. The evaluation is focused on small size dataset, since DES techniques has shown to be an effective tool for problems where the level of uncertainty for recognition is high due to few training samples~\cite{paulo2}. However, a few larger datasets were also considered in order to evaluate the performance of the proposed framework under different conditions. The goal of the experiments is to answer the following research questions: (1) Can the use of multiple DES criteria, as meta-features, lead to a more robust dynamic selection technique? (2) Does the proposed framework outperform current DES techniques for ill-defined problems?

This paper is organized as follows: Section~\ref{sec:relatedWork} introduces the notion of classifier competence, and the state-of-the-art techniques for dynamically measuring the classifiers' competence are presented. The proposed framework is presented in Section~\ref{sec:proposedMethod}. The experimental study is conducted in Section~\ref{sec:experiments}. Finally, our conclusion is presented in the last section.

\section{Classifier competence for dynamic selection}
\label{sec:relatedWork}

Classifier competence defines how much we trust an expert, given a classification task. The notion of competence used is extensively in the field of machine learning as a way of selecting, from the plethora of different classification models, the one that best fits the given problem. 
Let $C = \{c_{1}, \ldots, c_{M}\}$ ($M$ is the size of the pool of classifiers) be the pool of classifiers and $c_{i}$ a base classifier belonging to the pool $C$. The goal of dynamic selection is to find an ensemble of classifiers $C' \subset C$ that has the best classifiers to classify a given test sample $\mathbf{x}_{j}$. This is different from static selection, where the ensemble of classifiers $C'$ is selected during the training phase, and considering the global performance of the base classifiers over a validation dataset~\cite{GiacintoR01,Cruz2012,multiga,greedy}. 

Nevertheless, the key issue in dynamic selection is how to measure the competence of a base classifier $c_{i}$ for the classification of a given query sample $\mathbf{x}_{j}$. In the literature, we can observe three categories: the classifier accuracy over a local region, i.e., in a region of the feature space surrounding the query instance $\mathbf{x}_{j}$, decision templates~\cite{decisiontemplates}, which are techniques that work in the decision space (i.e, a space defined by the outputs of the base classifiers) and the extent of consensus or confidence. The three categories are described in the following subsections.

\subsection{Classifier accuracy over a local region}
\label{sec:localaccuracy}

Classifier accuracy is the most commonly used criterion for dynamic classifier and ensemble selection techniques~\cite{lca,knora,ijcnn2011,clussel,selectionfusion,singh,classrank,Smits_2002,dceid}. Techniques that are based on local accuracy first define a small region in the feature space surrounding a given test instance $\mathbf{x}_{j}$, called the region of competence. This region is computed using either the K-NN algorithm~\cite{knora,lca,ijcnn2011} or by Clustering techniques~\cite{clussel,selectionfusion}, and can be defined either in the training set~\cite{lca} or in the validation set, such as in the KNORA techniques~\cite{knora}.  

Based on the samples belonging to the region of competence, a criterion is applied in order to measure the level of competence of a base classifier.
For example, the Overall Local Accuracy (OLA)~\cite{lca} technique uses the accuracy of the base classifier in the whole region of competence as a criterion to measure its level of competence. The classifier that obtains the highest accuracy rate is considered the most competent one. The Local Classifier Accuracy (LCA)~\cite{lca} computes the performance of the base classifier in relation to a specific class label using a posteriori information~\cite{DidaciGRM05}. The Modified Local Accuracy~\cite{Smits_2002} works similarly to the LCA technique, with the only difference being that each sample belonging to the region of competence is weighted by its Euclidean distance to the query instance. That way, instances from the region of competence that are closer to the test sample have a higher influence when computing the performance of the base classifier. The classifier rank method~\cite{classrank} uses the number of consecutive correctly classified samples as a criterion to measure the level of competence. The classifier that correctly classifies the most consecutive samples coming from the region of competence is considered to have the highest competence level or ``rank''.

Ko et al.~\cite{knora} proposed the K-Nearest Oracles (KNORA) family of techniques, inspired by the Oracle concept. Four techniques are proposed: the KNORA-Eliminate (KNORA-E) which, considers that a base classifier $c_{i}$ is competent for the classification of the query instance $\mathbf{x}_{j}$ if $c_{i}$ achieves a perfect accuracy for the whole region of competence. Only the base classifiers with a perfect accuracy are used during the voting scheme. In The KNORA-Union (KNORA-U) technique, the level of competence of a base classifier $c_{i}$ is measured by the number of correctly classified samples in the defined region of competence. In this case, every classifier that correctly classified at least one sample can submit a vote. In addition, two weighted versions, KNORA-E-W and KNORA-U-W were also proposed, in which the influence of each sample belonging to the region of competence was weighted based on its Euclidean distance to the query sample $\mathbf{x}_{j}$. Lastly, Xiao et al.~\cite{dceid} proposed the Dynamic Classifier Ensemble for Imbalanced Data (DCEID), which is based on the same principles as the LCA technique. However, this technique also takes into account each class prior probability when computing the performance of the base classifier for the defined region of competence in order to deal with imbalanced distributions.

The difference between these techniques lies in how they utilize the local accuracy information in order to measure the level of competence of a base classifier. The main issue with the techniques arises from the fact that they depend on the performance of the techniques that define the region of competence such as K-NN or clustering techniques. In our previous work~\cite{ijcnn2011}, we demonstrated that the effectiveness of dynamic selection techniques is limited by the performance of the algorithm that defines the region of competence. The dynamic selection technique is likely to commit errors when outlier instances (i.e., mislabelled samples) exists around the query sample in the feature space~\cite{ijcnn2011}. Using the local accuracy information alone is not sufficient to achieve results close to the Oracle. Moreover, any difference between the distribution of validation and test datasets may negatively affect the system performance. Consequently, we believe that additional information should also be considered. 

\subsection{Decision Templates}
\label{sec:decisionTemplates}

In this class of methods, the goal is also to select samples that are close to the query instance $\mathbf{x}_{j}$. However, the similarity is computed over the decision space through the concept of decision templates~\cite{decisiontemplates}. This is performed by transforming both the test instance $\mathbf{x}_{j}$ and the validation data into output profiles. The output profile of an instance $\mathbf{x}_{j}$ is denoted by $\tilde{\mathbf{x}}_{j} = \left\lbrace \tilde{\mathbf{x}}_{j,1}, \tilde{\mathbf{x}}_{j,2}, \ldots, \tilde{\mathbf{x}}_{j,M} \right\rbrace $, where each $\tilde{\mathbf{x}}_{j,i}$ is the decision yielded by the base classifier $c_{i}$ for the sample $\mathbf{x}_{j}$.

Based on the information extracted from the decision space, the K-Nearest Output Profile (KNOP)~\cite{logid} is similar to the KNORA technique, with the difference being that the KNORA works in the feature space, while the KNOP works in the decision space. 
The KNOP technique first defines a set with the samples that are most similar to the output profile of the input sample, $\tilde{\mathbf{x}}_{j}$ in the decision space, called the output profiles set. The validation set is used for this purpose. Then, similarly to the KNORA-E technique, only the base classifiers that achieve a perfect recognition accuracy for the samples belonging to the output profiles set are used during the voting scheme.
The Multiple Classifier Behaviour (MCB) technique~\cite{mcb} also defines a set with the most similar output profiles to the input sample using the decision space. Here, the selection criterion is based on a threshold. The base classifiers that achieve a performance higher than the predefined threshold are considered competent and are selected to form the ensemble.

The advantage of this class of methods is that they are not limited by the quality of the region of competence defined in the feature space, with the similarity computed based on the decision space rather than the feature space. However, the disadvantage with this comes from the fact that only global information is considered, while the local expertise of each base classifier is neglected.

\subsection{Extent of Consensus or confidence}
\label{sec:confidence}

Different from other methods, techniques that are based on the extent of consensus work by considering a pool of ensemble of classifiers (EoC) rather than a pool of classifiers. Hence, the first step is to generate a population of EoC, $C^{*} = \{C^{'}_{1},C^{'}_{2}, \ldots, C^{'}_{M'}\}$ ($M^{'}$ is the number of EoC generated) using an optimization algorithm such as genetic algorithms or greedy search~\cite{multiga,docs,greedy}. Then, for each new query instance $\mathbf{x}_{j}$, the level of competence of an ensemble of classifiers $C^{'}_{i}$ is equal to the extent of consensus among its base classifiers. 

Several criterion based on this paradigm was proposed: the Margin-based Dynamic Selection (MDS)~\cite{docs}, where the criterion is the margin between the most voted class and the second most voted class. The margin is computed simply by considering the difference between the number of votes received by the most voted class and those received by the second most voted class. Two variations of the MDS where proposed in~\cite{docs}, the Class-Strength Dynamic Selection (CSDS), which includes the ensemble decision in the computation of the MDS, and the GSDS, where the global performance of each EoC is also taken into account~\cite{paulo2}. Another technique from this paradigm is the Ambiguity-guided Dynamic Selection (ADS)~\cite{docs}, which uses the ambiguity among the base classifiers of an EoC as the criterion for measuring the competence level of an EoC. The ambiguity is calculated by the number of base classifiers of an ensemble that disagrees with the ensemble decision. The lower the number of classifiers that disagree with the ensemble decision, the higher the level of competence of the EoC. 

The greatest advantage of this class of methods stems from the fact that it does not require information from the region of competence. Thus, it does not suffer from the limitations of the algorithm that defines the region of competence. However, these techniques present the following disadvantages: In many cases, the search cannot find an EoC with an acceptable confidence level. There is a tie between different members of the pool, and the systems end up performing a random decision~\cite{paulo2}. In addition, some classifiers are more overtrained than others. In this case, they end up dominating the outcome even though they do not present better recognition performance~\cite{trainnottrain}. The pre-computation of ensembles also greatly increases the overall system complexity as we are dealing with a pool of EoC rather than a pool of classifiers.

\section{The Proposed Framework: META-DES}
\label{sec:proposedMethod}

\subsection{Problem definition}

From the meta-learning perspective, the dynamic selection problem can be seen as another classification problem, called the meta-problem. This meta-problem uses different criteria regarding the behavior of a base classifier in order to decide whether it is competent enough to classify a given sample $\mathbf{x}_{j}$. Thus, a dynamic selection system can be defined based on two environments. A classification environment in which the input features are mapped into a set of class labels $w = \left \{  w_{1}, w_{2}, ..., w_{L}  \right \}$ and a meta-classification environment in which information about the behavior of the base classifier is extracted from the classification environment and used to decide whether a base classifier $c_{i}$ is competent enough to classify $\mathbf{x}_{j}$. 

To keep with the conventions of the meta-learning literature, we define the proposed dynamic ensemble selection in a meta-learning framework as follows:
 
 \begin{itemize}
 
 \item The \textbf{meta-problem} consists in defining whether a base classifier $c_{i}$ is competent enough to classify $\mathbf{x}_{j}$.
 
 \item The \textbf{meta-classes} of this meta-problem are either ``competent'' or ``incompetent'' to classify $\mathbf{x}_{j}$.
 
 \item Each \textbf{meta-feature} $f_{i}$ corresponds to a different criterion to measure the level of competence of a base classifier.
 
 \item The meta-features are encoded into a \textbf{meta-features vector} $v_{i,j}$ which contains the information about the behavior of a base classifier $c_{i}$ in relation to the input instance $\mathbf{x}_{j}$. 
 
 \item A \textbf{meta-classifier} $\lambda$ is trained based on the meta-features $v_{i,j}$ to predict whether or not $c_{i}$ will achieve the correct prediction for $\mathbf{x}_{j}$.
 
 \end{itemize}
 
In other words, a meta-classifier $\lambda$ is trained, based on $v_{i,j}$, to predict whether a base classifier $c_{i}$ is competent enough to classify given a test sample $\mathbf{x}_{j}$. Thus, the proposed system differs from the current state-of-the-art dynamic selection techniques not only because it uses multiple criteria, but also because the selection rule is learned by the meta-classifier $\lambda$ using the training data.

\subsection{The proposed META-DES}
\label{sec:proposed}

\begin{figure*}[!ht]
  
   \begin{center}  	 
       	  \epsfig{file=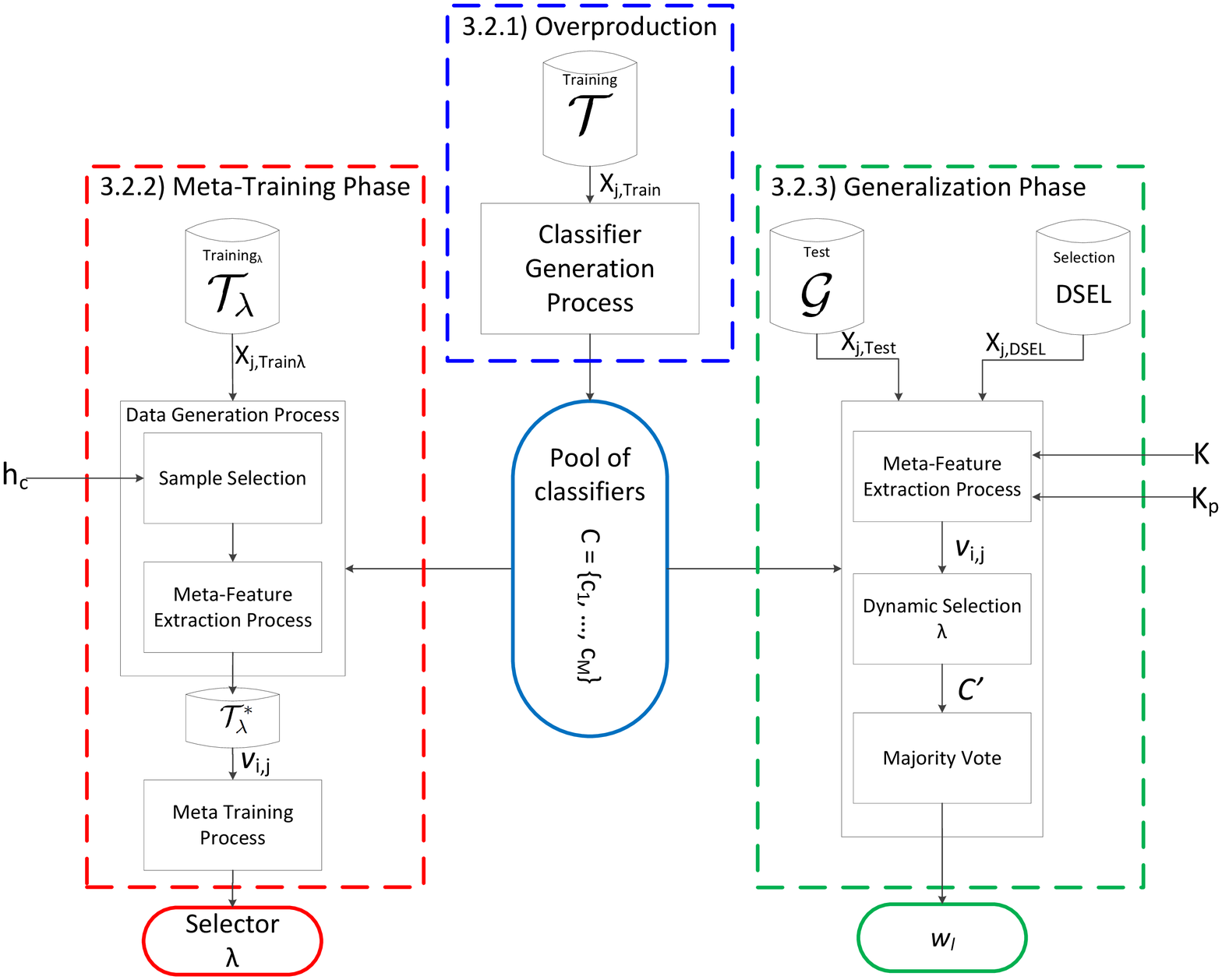, clip=,  width=1.0\textwidth}
   \end{center}
\caption{Overview of the proposed META-DES framework. It is divided into three steps 1) Overproduction, where the pool of classifiers $C = \{c_{1}, \ldots, c_{M}\}$ is generated, 2) The training of the meta-classifier $\lambda$, and 3) The generalization phase where an ensemble $C'$ is dynamically defined based on the meta-information extracted from $\mathbf{x}_{j,test}$ and the pool $C = \{c_{1}, \ldots, c_{M}\}$. The generalization phase returns the label $w_{l}$ of $\mathbf{x}_{j,test}$. $h_{C}$, $K$ and $K_{p}$ are the hyper-parameters required by the proposed system.}
\label{fig:overview}
\end{figure*}

The META-DES framework is divided into three phases (Figure~\ref{fig:overview}): 

\begin{enumerate}

\item  The overproduction phase, where the pool of classifiers $C = \{c_{1}, \ldots, c_{M}\}$, composed of $M$ classifiers, is generated using the training instances $\mathbf{x}_{j,train}$ from the dataset $\mathcal{T}$.

\item The meta-training stage, in which samples $\mathbf{x}_{j,train_{\lambda}}$ from the meta-training dataset $\mathcal{T}_{\lambda}$ are used to extract the meta-features. A different dataset $\mathcal{T}_{\lambda}$ is used in this phase in order to prevent overfitting. The meta-feature vectors $v_{i,j}$ are stored in the set $\mathcal{T}_{\lambda}^{*}$ that is later used to train the meta-classifier $\lambda$.

\item  The generalization phase, given a test sample $\mathbf{x}_{j,test}$ resulting from the generalization data $\mathcal{G}$; its region of competence is extracted using the samples from the dynamic selection dataset $D_{SEL}$ in order to compute the meta-features. The meta-feature vector $v_{i,j}$ is then passed to the selector $\lambda$, which decides whether $c_{i}$ is competent enough to classify $\mathbf{x}_{j,test}$ and should be added to the ensemble, $C'$. The majority vote rule is applied over the ensemble $C'$, giving the classification $w_{l}$ of $\mathbf{x}_{j,test}$.

\end{enumerate}

\subsubsection{Overproduction}

In this work, the Overproduction phase is performed using the Bagging technique~\cite{bagging,baggingfor}. Bagging is an acronym for Bootstrap AGGregatING. The idea behind this technique is to build a diverse ensemble of classifiers by randomly selecting different subsets of the training data. Each subset is used to train one individual classifier $c_{i}$. As the focus of the paper is on classifier selection, and not on classifier generation methods, only the bagging technique is considered.

\subsubsection{Meta-training}

As shown in Figure~\ref{fig:overview}, the meta-training stage consists of three steps: the sample selection process, the meta-features extraction process, and the training of the meta-classifier $\lambda$. For every sample $\mathbf{x}_{j,train_{\lambda}} \in \mathcal{T}_{\lambda}$, the first step is to apply the sample selection mechanism in order to know whether or not $\mathbf{x}_{j,train_{\lambda}}$ should be used for the training of the meta-classifier $\lambda$. The whole Meta-training phase is formalized in Algorithm~\ref{alg:trainingselector}.

\begin{algorithm}[!ht]
	\caption{The Meta-Training Phase}
	\label{alg:trainingselector}
		\begin{algorithmic}[1]
		
			\REQUIRE Training data $\mathcal{T_{\lambda}}$
			\REQUIRE Pool of classifiers $C = \{c_{1}, \ldots, c_{M}\}$
			\STATE $\mathcal{T}_{\lambda}^{*} = \emptyset$
			
			\FORALL{$\mathbf{x}_{j,train_{\lambda}} \in \mathcal{T_{\lambda}}$}
			
				\STATE Compute the consensus of the pool $H \left ( \mathbf{x}_{j,train_{\lambda}}, C \right )$
				\IF{$H \left ( \mathbf{x}_{j,train_{\lambda}}, C \right ) < h_{C}$}
			
					\STATE Find the region of competence $\theta_{j}$ of $\mathbf{x}_{j,train_{\lambda}}$ using $\mathcal{T}_{\lambda}$.
					\STATE Compute the output profile $\tilde{\mathbf{x}}_{j,train_{\lambda}}$ of $\mathbf{x}_{j,train_{\lambda}}$.
					\STATE Find the $K_{p}$ similar output profiles $\phi_{j}$ of $\tilde{\mathbf{x}}_{j,train_{\lambda}}$ using $\tilde{\mathcal{T}_{\lambda}}$.  
					
					\FORALL{$c_{i} \in C$}
					
						\STATE $v_{i,j} = MetaFeatureExtraction(\theta_{j}, \phi_{j}, c_{i}, \mathbf{x}_{j,train_{\lambda}})$
						\IF{$c_{i}$ correctly classifies $\mathbf{x}_{j,train_{\lambda}}$}
							\STATE  $\alpha_{i,j} = 1$ ``$c_{i}$ is competent for $\mathbf{x}_{j,train_{\lambda}}$''
						\ELSE
							\STATE  $\alpha_{i,j} = 0$  `` $c_{i}$ is incompetent for $\mathbf{x}_{j,train_{\lambda}}$''
						\ENDIF
						\STATE $\mathcal{T}_{\lambda}^{*} = \mathcal{T}_{\lambda}^{*} \cup \left\lbrace  v_{i,j} \right\rbrace$
					\ENDFOR
				\ENDIF
			\ENDFOR
			\STATE Divide $\mathcal{T}_{\lambda}^{*}$ into $25\%$ for validation and $75\%$ for training.
			\STATE Train $\lambda$ using the Levenberg-Marquadt algorithm. 
			\RETURN The meta-classifier $\lambda$.
		\end{algorithmic}
\end{algorithm}

\paragraph{Sample Selection}

As demonstrated by Dos Santos et al.~\cite{docs} and Cavalin et al.~\cite{paulo2}, one of the main issues in dynamic ensemble selection arises when classifying testing instances where the degree of consensus among the pool of classifier is low, i.e., when the number of votes from the winning class is close or even equal to the number of votes from the second class. To tackle this issue, we decided to focus the training of the meta-classifier $\lambda$ to specifically deal with cases where the extent of consensus among the pool is low. This step is conducted using a threshold $h_{C}$, called the consensus threshold. Each instance $\mathbf{x}_{j,train_{\lambda}}$ is first evaluated by the whole pool of classifiers in order to compute the degree of consensus among the pool, denoted by $H \left ( \mathbf{x}_{j,train_{\lambda}}, C \right )$. If the consensus $H \left ( \mathbf{x}_{j,train_{\lambda}}, C \right )$ falls below the consensus threshold $h_{C}$, the instance $\mathbf{x}_{j,train_{\lambda}}$ is used to compute the meta-features.


Before extracting the meta-features, the region of competence of the instance $\mathbf{x}_{j,train_{\lambda}}$, denoted by $\theta_{j} = \left \{ \mathbf{x}_{1}, \ldots, \mathbf{x}_{K} \right \}$, must first be computed. The region of competence $\theta_{j}$ is defined in the $\mathcal{T_{\lambda}}$ set, using the K-Nearest Neighbor algorithm (line 5). Then, $\mathbf{x}_{j,train_{\lambda}}$ is transformed into an output profile. The output profile of the instance $\mathbf{x}_{j,train_{\lambda}}$ is denoted by $\tilde{\mathbf{x}}_{j,train_{\lambda}} = \left\lbrace \tilde{\mathbf{x}}_{j,train_{\lambda},1}, \tilde{\mathbf{x}}_{j,train_{\lambda},2}, \ldots, \tilde{\mathbf{x}}_{j,train_{\lambda},M} \right\rbrace $, where each $\tilde{\mathbf{x}}_{j,train_{\lambda},i}$ is the decision yielded by the base classifier $c_{i}$ for the sample $\mathbf{x}_{j,train_{\lambda}}$~\cite{paulo2,logid}.

Next, with the region of competence $\theta_{j}$ and the set with the most similar output profiles $\phi_{j}$ computed, for each base classifier $c_{i}$ belonging to the pool of classifiers $C$, one meta-feature vector $v_{i,j}$ is extracted (lines 8 to 14). Each $v_{i,j}$ contains five sets of meta-features:

\paragraph{Meta-feature extraction process}
\label{sec:metafeatures}

Five different sets of meta-features are proposed in this work. Each feature set $f_{i}$, corresponds to a different criterion for measuring the level of competence of a base classifier. Each set captures a different property about the behavior of the base classifier, and can be seen as a different criterion to dynamically estimate the level of competence of base classifier such as, the classification performance estimated in a local region of the feature space and the classifier confidence for the classification of the input sample. 
Using five distinct sets of meta-features, even though one criterion might fail due to imprecisions in the local regions of the feature space or due to low confidence results, the system can still achieve a good performance as other meta-features are considered by the selection scheme. Table~\ref{table:metafeatures} shows the criterion used by each $f_{i}$ and its relationship with one dynamic ensemble selection paradigm presented in Section~\ref{sec:relatedWork}.


\begin{table}[htbp]
    \centering
    \caption{Relationship between each meta-features and different paradigms to compute the level of competence of a base classifier.}
     \label{table:metafeatures} 
     \begin{tabular}{|l | l | l|}
    \hline
       \textbf{Meta-Feature} & \textbf{Criterion} & \textbf{Paradigm}\\
        \hline
        $f_{1}$  & Local accuracy in the region of competence & Classifier Accuracy over a local region   \\
		\hline
        $f_{2}$ & Extent of consensus in the region of competence & Classifier consensus  \\
		\hline
        $f_{3}$  & Overall accuracy in the region of competence & Accuracy over a local region \\
		\hline
        $f_{4}$  & Accuracy in the decision space & Decision Templates\\
       \hline
        $f_{5}$  & Degree of confidence for the input sample & Classifier confidence\\
      \hline
      
    \end{tabular}
\end{table}

Three meta-features, $f_{1}$, $f_{2}$ and $f_{3}$, are computed using information extracted from the region of competence $\theta_{j}$. $f_{4}$ uses information extracted from the set of output profiles $\phi_{j}$. $f_{5}$ is calculated directly from the input sample $\mathbf{x}_{j,train_{\lambda}}$, and corresponds to the level of confidence of $c_{i}$ for the classification of $\mathbf{x}_{j,train_{\lambda}}$.

\begin{description}

\item  [\textbf{$f_{1}$ -} Neighbors' hard classification:] First, a vector with $K$ elements is created. For each instance $\mathbf{x}_{k}$, belonging to the region of competence $\theta_{j}$, if $c_{i}$ correctly classifies $\mathbf{x}_{k}$, the $k$-th position of the vector is set to 1, otherwise it is 0. Thus, $K$ meta-features are computed. 

\item [\textbf{$f_{2}$ -} Posterior probability:] First, a vector with $K$ elements is created. Then, for each instance $\mathbf{x}_{k}$, belonging to the region of competence $\theta_{j}$, the posterior probability of $c_{i}$, $P(w_{l}\mid \mathbf{x}_{k})$ is computed and inserted into the $k$-th position of the vector. Consequently, $K$ meta-features are computed. 

\item [\textbf{$f_{3}$ -} Overall Local accuracy:] The accuracy of $c_{i}$ over the whole region of competence $\theta_{j}$ is computed and encoded as $f_{3}$. 

\item [\textbf{$f_{4}$ -} Output profiles classification:] First, a vector with $K_{p}$ elements is generated. Then, for each member $\tilde{\mathbf{x}}_{k}$ belonging to the set of output profiles $\phi_{j}$, if the label produced by $c_{i}$ for $\mathbf{x}_{k}$ is equal to the label $w_{l,k}$ of $\tilde{\mathbf{x}}_{k}$, the $k$-th position of the vector is set to 1, otherwise it is 0. A total of $K_{p}$ meta-features are extracted using output profiles.

\item [\textbf{$f_{5}$ -} Classifier's Confidence:] The perpendicular distance between the input sample $\mathbf{x}_{j,train_{\lambda}}$ and the decision boundary of the base classifier $c_{i}$ is calculated and encoded as $f_{5}$. $f_{5}$ is normalized to a $[0-1]$ range using the Min-max normalization.

\end{description}

\begin{figure}[htbp]
    \begin{center}
    	  \epsfig{file=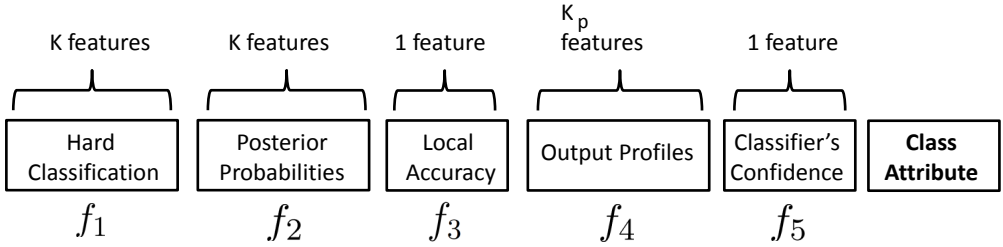, clip=, width=0.80\textwidth}
    	  \end{center}
          \caption{Feature Vector containing the meta-information about the behavior of a base classifier. A total of 5 different meta-features are considered. The size of the feature vector is $(2 \times K) + K_{p} + 2$. The class attribute indicates whether or not $c_{i}$ correctly classified the input sample.}
    \label{fig:featVector}
\end{figure}

A vector $v_{i,j} = \left\lbrace f_{1} \cup f_{2} \cup f_{3} \cup f_{4} \cup f_{5} \right\rbrace$ is obtained at the end of the process (Figure~\ref{fig:featVector}). If $c_{i}$ correctly classifies $\mathbf{x}_{j,train_{\lambda}}$, the class attribute of $v_{i,j}$, $\alpha_{i,j} = 1$ (i.e., $v_{i,j}$ corresponds to the behavior of a competent classifier), otherwise $\alpha_{i,j} = 0$. $v_{i,j}$ is stored in the meta-features dataset $\mathcal{T}_{\lambda}^{*}$ (lines 10 to 16). 

For each sample $\mathbf{x}_{j,train_{\lambda}}$ used in the meta-training stage, a total of $M$ ($M$ is the size of the pool of classifiers $C$) meta-feature vectors $v_{i,j}$ are extracted, each one corresponding to one classifier from the pool $C$. In this way, the size of the meta-training dataset $\mathcal{T}_{\lambda}^{*}$ is the pool size $M \times$ number of training samples $N$. For instance, consider that $200$ training samples are available for the meta-training stage ($N = 200$), if the pool $C$ is composed of $100$ weak classifiers ($M = 100$), the meta-training dataset is the number of training samples $N$ $\times$ the number classifiers in the pool $M$, $N * M = 20.000$. 
Hence, even though the classification problem may be ill-defined due to the size of the training set, we can overcome this limitation in the meta-problem by increasing the size of the pool of classifiers.

\paragraph{Training}

The last step of the meta-training phase is the training of the meta-classifier $\lambda$. The dataset $\mathcal{T}_{\lambda}^{*}$ is divided on the basis of 75\% for training and 25\% for validation. A Multi-Layer Perceptron (MLP) neural network is considered as the selector $\lambda$. The validation data was used to select the number of nodes in the hidden layer. We use a configuration of 10 neurons in the hidden layer since there were no improvement in results with more than 10 neurons. The training process for $\lambda$ is performed using the Levenberg-Marquadt algorithm. In addition, the training process is stopped if its performance on the validation set decreases or fails to improve for five consecutive epochs.

\subsubsection{Generalization Phase}
\label{sec:dynamicSelection}

\begin{algorithm}[htbp]
	\caption{Classification steps using the selector $\lambda$}
	\label{alg:generalization}
		\begin{algorithmic}[1]
		
			\REQUIRE Query sample $\mathbf{x}_{j,test}$
			\REQUIRE Pool of classifiers $C = \{c_{1}, \ldots, c_{M}\}$
			\REQUIRE dynamic selection dataset $D_{SEL}$
			\STATE $C^{'} = \emptyset$
			
			\STATE Find the region of competence $\theta_{j}$ of $\mathbf{x}_{j,test}$ using $D_{SEL}$.
			\STATE Compute the output profile $\tilde{\mathbf{x}}_{j,test}$  of  $\mathbf{x}_{j,test}$.
			\STATE Find the $K_{p}$ similar output profiles $\phi_{j}$ of $\tilde{\mathbf{x}}_{j,test}$ using $\tilde{D}_{SEL}$.  
			
			\FORALL{$\mathbf{c}_{i} \in C$}
					
					\STATE $v_{i,j} = FeatureExtraction(\theta_{j},\phi_{j}, c_{i}, \mathbf{x}_{j,test})$ 
					\STATE input $v_{i,j}$ to $\lambda$
				
				\IF{$\alpha_{i,j} = 1$ ``$c_{i}$ is competent for $\mathbf{x}_{j,test}$''}
				
					\STATE $C^{'} = C^{'} \cup \left\lbrace  c_{i} \right\rbrace$
				
				\ENDIF
			\ENDFOR

			\STATE $w_{l} = MajorityVote(\mathbf{x}_{j,test},C^{'})$
			\RETURN $w_{l}$
			
		\end{algorithmic}
\end{algorithm}

The generalization procedure is formalized by Algorithm~\ref{alg:generalization}. Given the query sample $\mathbf{x}_{j,test}$, in this phase, the region of competence $\theta_{j}$ is computed using the samples from the dynamic selection dataset $D_{SEL}$ (line 2). Following that, the output profiles $\tilde{\mathbf{x}}_{j,test}$ of the test sample, $\mathbf{x}_{j,test}$, are calculated. The set with $K_{p}$ similar output profiles $\phi_{j}$, of the query sample $\mathbf{x}_{j,test}$, is obtained through the Euclidean distance applied over the output profiles of the dynamic selection dataset, $\tilde{D}_{SEL}$.

Next, for each classifier $c_{i}$ belonging to the pool of classifiers $C$, the meta-feature extraction process is called (Section 3.2.2.2), returning the meta-features vector $v_{i,j}$ (lines 5 and 6). Then, $v_{i,j}$ is used as input to the meta-classifier $\lambda$. If the output of $\lambda$ is $1$ (i.e., competent), $c_{i}$ is included in the ensemble $C'$ (lines 8 to 10). After every base classifier, $c_{i}$, is evaluated, the ensemble $C'$ is obtained. The base classifiers in $C'$ are combined through the Majority Vote rule~\cite{kittler}, giving the label $w_{l}$ of $x_{j,test}$ (line 12 and 13). The majority vote rule is used to combine the selected classifiers since it has been successfully used by other DES techniques~\cite{Alceu2014}. Tie-breaking is handled by choosing the class with the highest a posteriori probability.

\section{Experiments}
\label{sec:experiments}

\subsection{Datasets}

A total of 30 datasets are used in the comparative experiments. sixteen coming from the UCI machine learning repository~\cite{Lichman2013}, four from the STATLOG project~\cite{King95statlog}, four from the Knowledge Extraction based on Evolutionary Learning (KEEL) repository~\cite{FdezFLDG11}, four from the Ludmila Kuncheva Collection of real medical data~\cite{lkc}, and two artificial datasets generated with the Matlab PRTOOLS toolbox~\cite{PRTools}. We consider both ill-defined problems, such as, Heart and Liver Disorders as well as larger databases, such as, Adult, Magic Gamma Telescope, Phoneme and WDG V1. The key features of each dataset are shown in Table~\ref{table:datasets}.

\begin{table}[htbp]
    \centering
    \caption{Key Features of the datasets used in the experiments.}
     \label{table:datasets} 
     \resizebox{0.75\textwidth}{!}{
    \begin{tabular}{|c| c| c| c| c|}
    \hline
      \textbf{Database} & \textbf{ No. of Instances} & \textbf{Dimensionality} & \textbf{No. of Classes} & \textbf{Source}\\
        \hline

        \textbf{Pima} & 768 & 8 & 2 & UCI \\
		\hline	

        \textbf{Liver Disorders} & 345 & 6 & 2 & UCI  \\
		\hline

        \textbf{Breast (WDBC)} & 568 & 30 & 2 & UCI \\
		\hline

        \textbf{Blood transfusion} & 748 & 4 &	2 & UCI  \\
       	\hline 

	 \textbf{Banana}  & 1000 & 2 &	2 & PRTOOLS  \\
		\hline

        \textbf{Vehicle} & 846 & 18 & 4 & STATLOG \\
  		\hline

	\textbf{Lithuanian}  & 1000 & 2 & 2 & PRTOOLS  \\
      \hline 

        \textbf{Sonar} & 208 &	60 & 2 & UCI \\
        \hline

        \textbf{Ionosphere} & 315 &	34 & 2 & UCI  \\
        \hline

        \textbf{Wine} & 178 & 13 & 3 & UCI  \\
        \hline

        \textbf{Haberman's Survival} & 306 & 3 & 2 & UCI  \\
        \hline

       \textbf{Cardiotocography (CTG)} & 2126 & 21 & 3 & UCI \\    
       \hline

       \textbf{Vertebral Column} & 310 & 6 & 2 & UCI  \\          
       \hline

       \textbf{Steel Plate Faults} & 1941 & 27 & 7 & UCI \\   
       \hline

       \textbf{WDG V1} & 50000 & 21 & 3 & UCI  \\    
       \hline

       \textbf{Ecoli} & 336 & 7 & 8 & UCI   \\    
       \hline
       
	   
       \textbf{Glass} & 214 & 9 & 6  & UCI  \\                       
		\hline
		
       \textbf{ILPD} & 214 & 9 & 6  & UCI  \\                       
		\hline

        \textbf{Adult} & 48842 & 14 & 2 & UCI \\        
        \hline

       \textbf{Weaning} & 302 & 17 & 2 & LKC \\
        \hline

        \textbf{Laryngeal1} & 213 & 16 & 2 & LKC  \\        
       \hline         

       \textbf{Laryngeal3} & 353 & 16 & 3 & LKC \\    
       \hline

       \textbf{Thyroid} &  215 & 5 & 3 & LKC  \\
       \hline
      
        \textbf{German credit} & 1000 & 20 &2  & STATLOG  \\
        \hline

        \textbf{Heart} & 270 & 13  & 2  & STATLOG  \\
        \hline
        
        \textbf{Satimage} & 6435 & 19 & 7 & STATLOG   \\    
        \hline  

        \textbf{Phoneme} & 5404 & 6 & 2 & ELENA \\   
		\hline
		
		\textbf{Monk2}  & 4322 & 6 & 2 & KEEL  \\
		\hline    
		          
		\textbf{Mammographic}  & 961 & 5 & 2 & KEEL  \\
		\hline              
		          
    	\textbf{MAGIC Gamma Telescope}  & 19020 & 10 & 2 & KEEL  \\
       \hline

    \end{tabular}
		}
\end{table} 

\subsection{Experimental Protocol}

The experiments were conducted using 20 replications. For each replication, the datasets were randomly divided on the basis 50\% for training, 25\% for the dynamic selection dataset (D$_{SEL}$), and 25\% for the test set ($\mathcal{G}$). The divisions were performed maintaining the priors probabilities of each class. For the proposed META-DES, 50\% of the training data was used in the meta-training process $\mathcal{T}_{\lambda}$ and 50\% for the generation of the pool of classifiers ($\mathcal{T}$). 

For the two-class classification problems, the pool of classifiers was composed of 100 Perceptrons generated using the bagging technique~\cite{bagging}. For the multi-class problems, the pool of classifiers was composed of 100 multi-class perceptron classifier. The use of Perceptron as base classifier comes from the following observations based on past works in the literature:

\begin{itemize}

\item The use of weak classifiers can show more differences between the DES schemes~\cite{knora}. Thus, making it a better option for comparing different DES techniques.

\item Past works in the DES literature demonstrate that the use of weak models as base classifier achieve better results~\cite{docs,paulo2,outprof,eulanda,ijcnn2011}, where the use of decision trees or Perceptrons outperform strong classification models such as KNN classifiers.

\item As reported by Leo Breiman~\cite{bagging,baggingfor}, the bagging technique achieves better results when weak and unstable base classifiers are used.

\end{itemize}
 
\subsection{Parameters Setting}

The performance of the proposed selection scheme depends on three parameters: the neighborhood size, $K$, the number of similar patterns using output profiles $K_{p}$ and the consensus threshold $h_{C}$. The dynamic selection dataset $D_{SEL}$ was used for the analysis. The following methodology is used: 

\begin{itemize}
\item For the sake of simplicity, we selected the parameters that performed best. 
\item The value of the parameter $K$ was selected based on the results of our previous paper~\cite{ijcnn2011}. In this case, $K = 7$ showed the best overall results, considering several dynamic selection techniques. 
\item The Kruskall-Wallis statistical test with a 95\% confidence interval was used to determine whether the difference in results was statistically significant. If two configurations yielded similar results, we selected the one with the smaller parameter value as it leads to a smaller meta-features vector.
\item The parameter $h_{C}$ was evaluated with $K_{p}$ initially set at $1$. 
\item The best value of $h_{c}$ was used in the evaluation of the best value for $K_{p}$.
\item Only a subset with eleven of the thirty datasets are used for parameters setting procedure: Pima, Liver, Breast, Blood Transfusion, Banana, Vehicle, Lithuanian, Sonar, Ionosphere, Wine, Haberman's Survival.
 \end{itemize}

\subsubsection{The effect of the parameter $h_{C}$}

\begin{figure}[H]
    \begin{center}
    	  \epsfig{file=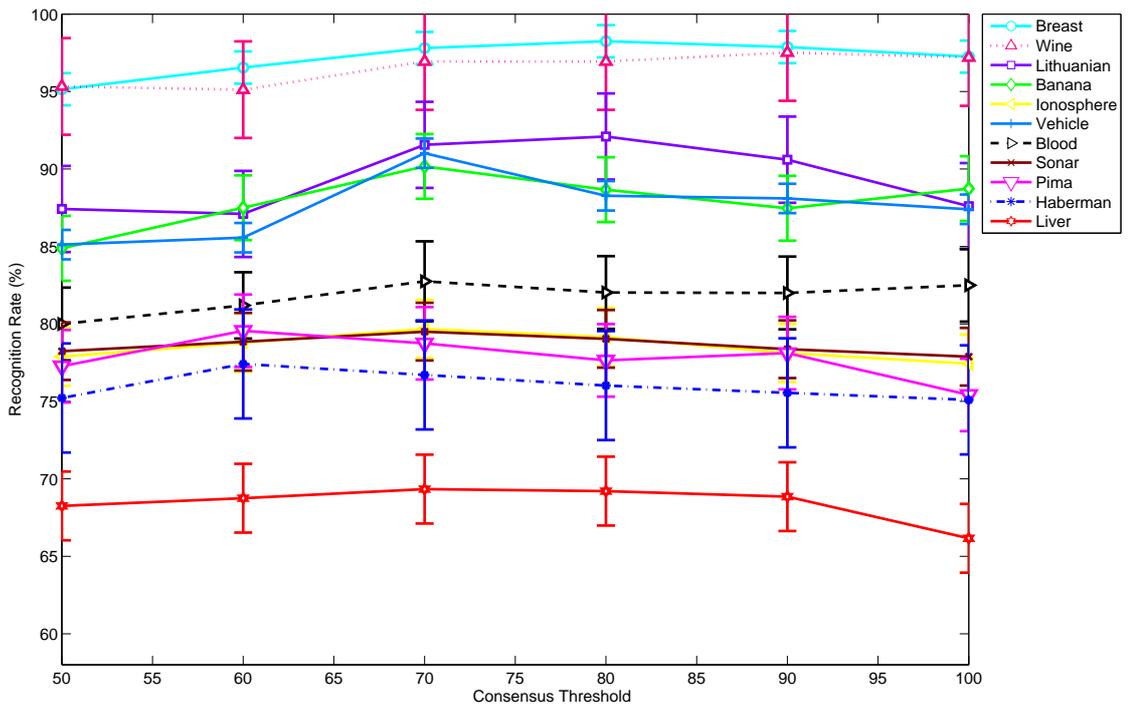, clip=, width=1.0\textwidth}
    	  \end{center}
          \caption{Performance of the proposed system based on the parameter $h_{C}$ on the dynamic selection dataset, $D_{SEL}$. $K = 7$ and $K_{p} = 1$.}
    \label{fig:percClass}
\end{figure}

We varied the parameter $h_{c}$ from 50\% to 100\% at 10 percentile point interval. Figure~\ref{fig:percClass} shows the mean performance and standard deviation for each $h_{C}$ value. We compared each pair of results using the Kruskal-Wallis non-parametric statistical test with a 95\% confidence interval. For 6 out of 11 datasets (Vehicle, Lithuanian, Banana, Blood transfusion, Ionosphere and Sonar) $h_{C} = 70\%$ presented a value that was statistically superior to the others. Hence, $h_{C} = 70\%$ was selected.

\subsubsection{The effect of the parameter $K_{p}$}

\begin{figure}[H]
    \begin{center}
    	  \epsfig{file=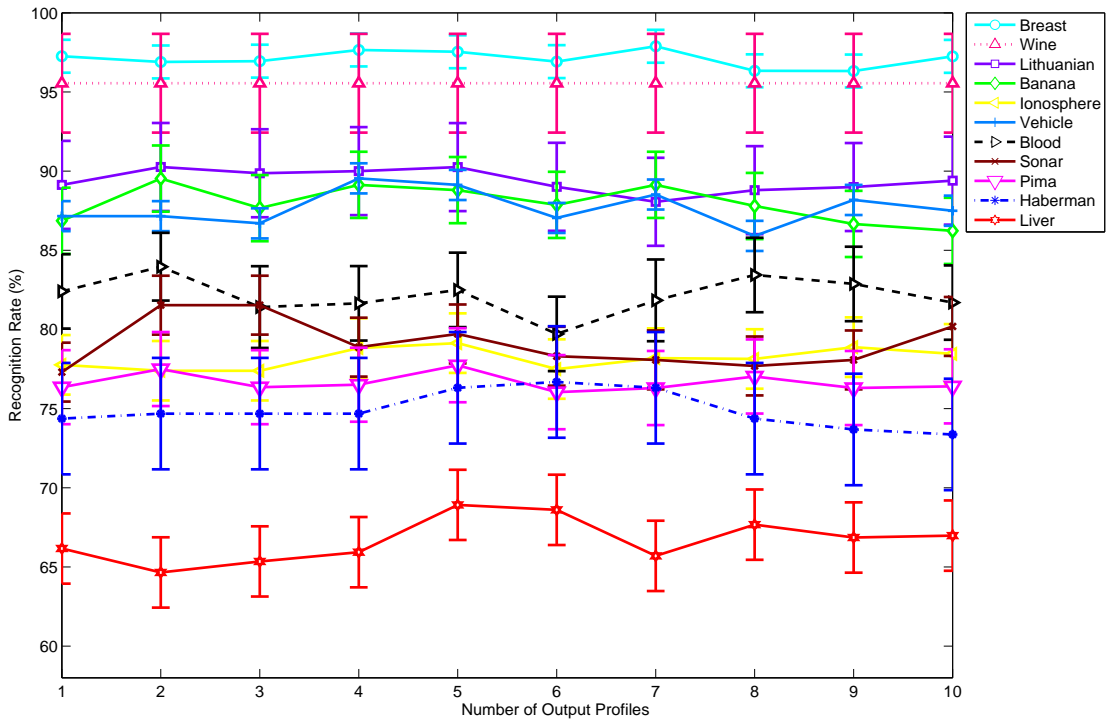, clip=, width=1.0\textwidth}
    	  \end{center}
          \caption{The performance of the system varying the parameter $K_{p}$ from 1 to 10 on the dynamic selection dataset, $D_{SEL}$. $h_{c} = 70\%$ and $K = 7$}
    \label{fig:difn}
\end{figure}

Figure~\ref{fig:difn} shows the impact of the value of the parameter $K_{p}$ in an 1-to-10 range. Once again, we compared each pair of results using the Kruskal-Wallis non-parametric statistical test, with a 95\% confidence. The results were statistically different only for the Sonar, Ionosphere and liver disorders datasets, where the value of $K_{p} = 5$ showed the best results. Hence, $K_{p}$ was set at $5$. 

\subsection{Comparison with the state-of-the-art dynamic selection techniques}

In this section we compare the recognition rates obtained by the proposed META-DES, against eight dynamic selection techniques found in the literature~\cite{Alceu2014}. The objective of this comparative study is to answer the following research question: (1) Can the use of multiple DES criteria as meta-features lead to a more robust dynamic selection technique? (2) Does the proposed framework outperform current DES techniques for ill-defined problems?

The eight state-of-the-art DES techniques used in this study are: the KNORA-ELIMINATE~\cite{knora}, KNORA-UNION~\cite{knora}, DES-FA~\cite{ijcnn2011}, Local Classifier Accuracy (LCA)~\cite{lca}, Overall Local Accuracy (OLA)~\cite{lca}, Modified Local Accuracy (MLA)~\cite{Smits_2002}, Multiple Classifier Behaviour (MCB)~\cite{mcb} and K-Nearests Output Profiles (KNOP)~\cite{logid,paulo2}. These techniques were selected because they presented the very best results in the dynamic selection literature according to a recent survey on this topic~\cite{Alceu2014}. In addition, we also compare the performance of the proposed META-DES with static combination methods (Adaboost and Bagging), the classifier with the highest accuracy in the validation data (Single Best), static ensemble selection based on the majority voting error~\cite{classmaj} and the abstract model (Oracle)~\cite{Kuncheva:2002}. The Oracle represents the ideal classifier selection scheme. It always selects the classifier that predicted the correct label, for any given query sample, if such classifier exists. For the static ensemble selection method, 50\% of the classifiers of the pool are selected. The comparison against static methods is used since it is suggested the DES literature that the minimum requirement for a DES method is to surpass the performance of static selection and combination methods in the same pool~\cite{Alceu2014}.

For all techniques, the pool of classifiers $C$ is composed of $100$ Perceptrons as base classifier ($M = 100$). For the state-of-the-art DES techniques (KNORA-E, KNORA-U, DES-FA, LCA, OLA, MLA, MCB and KNOP), the size of the region of competence (neighborhood size), $K$ is set to $7$, since it achieved the best result on previous publications~\cite{Alceu2014,ijcnn2011}. The size of the region of competence $K$ is the only hyper-parameter required for the eight DES techniques. For the Adaboost and Bagging technique $100$ iterations are used (i.e., $100$ base classifier are generated).

\begin{table*}[ht]
    \centering
    \caption{Mean and standard deviation results of the accuracy obtained for the proposed DES$_{D}$ and the DES systems in the literature. A pool of $100$ Perceptrons as base classifiers is used for all techniques. The best results are in bold. Results that are significantly better ($p < 0.05$) are marked with a $\bullet$.}
     \label{table:Results} 
     \resizebox{1.0\textwidth}{!}{
     \begin{tabular}{| c | c | c | c | c | c | c | c | c | c |}
    \hline

      \textbf{Database} & \textbf{META-DES} & \textbf{KNORA-E}~\cite{knora} & \textbf{KNORA-U}~\cite{knora} & \textbf{DES-FA}~\cite{ijcnn2011} & \textbf{LCA}~\cite{lca} & \textbf{OLA}~\cite{lca} & \textbf{MLA}~\cite{Smits_2002} & \textbf{MCB}~\cite{mcb} &  \textbf{KNOP}~\cite{paulo2} \\
        \hline

         \textbf{Pima} & \textbf{79.03(2.24)} $\bullet$ &  73.79(1.86) &  76.60(2.18) & 73.95(1.61) & 73.95(2.98)  & 73.95(2.56) & 77.08(4.56) & 76.56(3.71) & 73.42(2.11) \\
\hline
        \textbf{Liver Disorders} & \textbf{70.08(3.49)} $\bullet$ &  56.65(3.28) &  56.97(3.76) & 61.62(3.81) & 58.13(4.01)  & 58.13(3.27) & 58.00(4.25)  & 58.00(4.25) & 65.23(2.29) \\
\hline 
        \textbf{Breast (WDBC)} & 97.41(1.07) & 97.59(1.10)  &  97.18(1.02) & \textbf{97.88(0.78)} & 97.88(1.58)  & 97.88(1.58)  & 95.77(2.38)  & 97.18(1.38) & 95.42(0.89) \\
\hline 
        \textbf{Blood Transfusion} & \textbf{79.14(1.03)} $\bullet$ &  77.65(3.62) &  77.12(3.36) & 73.40(1.16) & 75.00(2.87) & 75.00(2.36) & 76.06(2.68)  & 73.40(4.19) & 77.54(2.03) \\
\hline      
        \textbf{Banana} & 91.78(2.68) &  93.08(1.67) & 92.28(2.87) & \textbf{95.21(3.18)} & 95.21(2.15)  & 95.21(2.15)  & 80.31(7.20) &  88.29(3.38) & 90.73(3.45) \\
\hline
        \textbf{Vehicle} & 82.75(1.70) &  83.01(1.54) &  82.54(1.70)  & 82.54(4.05) & 80.33(1.84)  & 81.50(3.24) & 74.05(6.65) & \textbf{84.90(2.01)} & 80.09(1.47) \\
\hline
        \textbf{Lithuanian Classes} & 93.18(1.32) &  93.33(2.50) &  95.33(2.64) & 98.00(2.46) & 85.71(2.20) & \textbf{98.66(3.85)} & 88.33(3.89)  & 86.00(3.33) & 89.33(2.29) \\
\hline        
        \textbf{Sonar} & \textbf{80.55(5.39)} &  74.95(2.79) &  76.69(1.94) & 78.52(3.86) & 76.51(2.06) & 74.52(1.54) & 76.91(3.20)  & 76.56(2.58) & 75.72(2.82) \\
\hline        
        \textbf{Ionosphere} & \textbf{89.94(1.96)} &  89.77(3.07) &  87.50(1.67) & 88.63(2.12) & 88.00(1.98)  & 88.63(1.98) & 81.81(2.52)  & 87.50(2.15) & 85.71(5.52) \\
\hline        
        \textbf{Wine} & \textbf{99.25(1.11)} $\bullet$ &  97.77(1.53) &  97.77(1.62) &  95.55(1.77)  & 85.71(2.25)  & 88.88(3.02) & 88.88(3.02)  & 97.77(1.62) & 95.50(4.14)\\
\hline        
        \textbf{Haberman} & \textbf{76.71(1.86)} &  71.23(4.16)  & 73.68(2.27)  & 72.36(2.41)  &  70.16(3.56) &  69.73(4.17) &  73.68(3.61)  &  67.10(7.65) & 75.00(3.40) \\        
 \hline
      \textbf{Cardiotocography (CTG)} & 84.62(1.08) & \textbf{86.27(1.57)} &  85.71(2.20) & \textbf{86.27(1.57)} &  86.65(2.35) &  86.65(2.35)  & 86.27(1.78) & 85.71(2.21) & 86.02(3.04) \\		
\hline
       \textbf{Vertebral Column} & 86.89(2.46) &  85.89(2.27) & \textbf{87.17(2.24)} &  82.05(3.20) &  85.00(3.25) &  85.89(3.74) &  77.94(5.80) & 84.61(3.95) & 86.98(3.21)\\		
\hline 
       \textbf{Steel Plate Faults} & 67.21(1.20) & 67.35(2.01) & 67.96(1.98)  & 68.17(1.59)  &  66.00(1.69) &  66.52(1.65)  & 67.76(1.54)  & 68.17(1.59) & \textbf{68.57(1.85)}   \\	
\hline
       \textbf{WDG V1} & \textbf{84.56(0.36)} &  84.01(1.10) &  84.01(1.10) & 84.01(1.10) & 80.50(0.56)  & 80.50(0.56) & 79.95(0.85)  & 78.75(1.35) & 84.21(0.45) \\	
\hline 
       \textbf{Ecoli} &  77.25(3.52) &  76.47(2.76)  & 75.29(3.41) & 75.29(3.41) &  75.29(3.41)  &  75.29(3.41) &  76.47(3.06) & 76.47(3.06) & \textbf{80.00(4.25)} $\bullet$ \\	
\hline	   
       \textbf{Glass} & 66.87(2.99) &  57.65(5.85) &  61.00(2.88) & 55.32(4.98) & 59.45(2.65) & 57.60(3.65) & 57.60(3.65) & \textbf{67.92(3.24)} & 62.45(3.65)\\	          
\hline		
       \textbf{ILPD} & 69.40(1.64) &  67.12(2.35) &  69.17(1.58) &  67.12(2.35) & \textbf{69.86(2.20)} & \textbf{69.86(2.20)} & \textbf{69.86(2.20)} &  68.49(3.27) & 68.49(3.27)  \\	     
\hline
        \textbf{Adult} & \textbf{87.15(2.43)} $\bullet$ &  80.34(1.57) &  79.76(2.26) & 80.34(1.57) & 83.58(2.32) &  82.08(2.42) & 80.34(1.32)  & 78.61(3.32) & 79.76(2.26) \\	 
\hline
       \textbf{Weaning} & \textbf{87.15(2.43)} $\bullet$ & 78.94(1.25) & 81.57(3.65) & 82.89(3.52) & 77.63(2.35) & 77.63(2.35) & 80.26(1.52) & 81.57(2.86) & 82.57(3.33) \\	
\hline
        \textbf{Laryngeal1} & \textbf{79.67(3.78)} $\bullet$ &  77.35(4.45) & 77.35(4.45) &  77.35(4.45) &  77.35(4.45) & 77.35(4.45) &  75.47(5.55) & 77.35(4.45) & 77.35(4.45) \\	
\hline
       \textbf{Laryngeal3} & 72.65(2.17) & 70.78(3.68) & 72.03(1.89) & 72.03(1.89) &  72.90(2.30) & 71.91(1.01) & 61.79(7.80) & 71.91(1.01) & \textbf{73.03(1.89)}  \\
\hline
       \textbf{Thyroid} & \textbf{96.78(0.87)} & 95.95(1.25) &  95.95(1.25) &  95.37(2.02) & 95.95(1.25) & 95.95(1.25) &  94.79(2.30) &  95.95(1.25) & 95.95(1.25) \\	
\hline
        \textbf{German credit} & \textbf{75.55(1.31)} $\bullet$ & 72.80(1.95) & 72.40(1.80)  & 74.00(3.30) & 73.33(2.85) & 71.20(2.52) & 71.20(2.52) & 73.60(3.30)  & 73.60(3.30) \\	
\hline
        \textbf{Heart} & 84.80(3.36) &  83.82(4.05) & 83.82(4.05) & 83.82(4.05) &  85.29(3.69) & 85.29(3.69) & \textbf{86.76(5.50)} & 83.82(4.05) & 83.82(4.05) \\	
\hline        
        \textbf{Satimage} & \textbf{96.21(0.87)} &  95.35(1.23) & 95.86(1.07) & 93.00(2.90) & 95.00(1.40) & 94.14(1.07) & 93.28(2.10) & 95.86(1.07) & 95.86(1.07) \\	
\hline
        \textbf{Phoneme} & \textbf{80.35(2.58)} &  79.06(2.50) &  78.92(3.33) &  79.06(2.50) & 78.84(2.53) &  78.84(2.53) &  64.94(7.75) & 73.37(5.55) & 78.92(3.33) \\	
\hline		
		\textbf{Monk2} & \textbf{83.24(2.19)} $\bullet$ & 80.55(3.32) & 77.77(4.25) & 75.92(4.25)  & 74.07(6.60) & 74.07(6.60) & 75.92(5.65) & 74.07(6.60) & 80.55(3.32) \\	
\hline		          
		\textbf{Mammographic} & \textbf{84.82(1.55)} $\bullet$ & 82.21(2.27) & 82.21(2.27) & 80.28(3.02) & 82.21(2.27 & 82.21(2.27) & 75.55(5.50) & 81.25(2.07) & 82.21(2.27) \\	
\hline				          
    	\textbf{MAGIC Gamma Telescope} & \textbf{84.35(3.27)} $\bullet$ &  80,03(3.25) & 79,99(3.55) & 81.73(3.27) & 81,53(3.35) & 81,16(3.00) & 73,13(6.35)  & 75,91(5.35) & 80,03(3.25) \\	
     
    \hline
    \end{tabular}
    }
 
\end{table*}

We split the results in two tables: Table~\ref{table:Results} shows a comparison with the proposed META-DES against the eight state-of-the-art dynamic selection techniques considered. A comparison of the META-DES against static combination rules is shown in Table~\ref{table:Results2}. Each pair of results is compared using the Kruskal-Wallis non-parametric statistical test, with a 95\% confidence interval. The best results are in bold. Results that are significantly better ($p < 0.05$) are marked with a $\bullet$. 

\begin{table}[!ht]
    \centering
    \caption{Mean and standard deviation results of the accuracy obtained for the proposed DES$_{D}$ and static ensemble combination. A pool of $100$ Perceptrons as base classifier is used for all techniques The best results are in bold. Results that are significantly better ($p < 0.05$) are marked with a $\bullet$.}
     \label{table:Results2} 
     \resizebox{0.73\textwidth}{!}{
     \begin{tabular}{|c |c |c |c |c |c||c|}
    \hline

      \textbf{Database} & \textbf{META-DES} & \textbf{Single Best~\cite{Alceu2014}} &  \textbf{Bagging}~\cite{bagging} & \textbf{AdaBoost}~\cite{boosting} & \textbf{Static Selection~\cite{classmaj}} & \textbf{Oracle}~\cite{Kuncheva:2002} \\
        \hline

        \textbf{Pima} & \textbf{79.03(2.24)} $\bullet$ & 73.57(1.49) & 73.28(2.08) & 72.52(2.48) & 72.86(4.78) & 95.10(1.19)  \\
\hline
        \textbf{Liver Disorders} & \textbf{70.08(3.49)} $\bullet$ & 65.38(3.47)  & 62.76(4.81) & 64.65(3.26) & 59.18(7.02) & 93.07(2.41) \\
 \hline
        \textbf{Breast (WDBC)} & 97.41(1.07) & 97.04(0.74) & 96.35(1.14) & \textbf{98.24(0.89)}  & 96.83(1.00) & 99.13(0.52) \\
 \hline
        \textbf{Blood Transfusion} & \textbf{79.14(1.03)} $\bullet$ & 75.07(1.83) & 75.24(1.67) & 75.18(2.08) & 75.74(2.23) & 94.20(2.08) \\
 \hline     
        \textbf{Banana} & \textbf{91.78(2.68)}  & 84.07(2.22) & 81.43(3.92) & 81.61(2.42) & 81.35(4.28) & 94.75(2.09)\\
\hline
        \textbf{Vehicle} & \textbf{82.75(1.70)}  & 81.87(1.47) & 82.18(1.31) & 80.56(4.51) & 81.65(1.48) & 96.80(0.94)\\
\hline
        \textbf{Lithuanian Classes} & \textbf{93.18(1.32)} $\bullet$ & 84.35(2.04) & 82.33(4.81) & 82.70(4.55) & 82.66(2.45) & 98.35 (0.57) \\
 \hline       
        \textbf{Sonar} & \textbf{80.55(5.39)}  & 78.21(2.36) & 76.66(2.36) & 74.95(5.21) & 79.03(6.50) & 94.46(1.63) \\
 \hline       
        \textbf{Ionosphere} & \textbf{89.94(1.96)}  & 87.29(2.28) & 86.75(2.75) & 86.75(2.34) & 87.50(2.23) & 96.20(1.72) \\
 \hline       
        \textbf{Wine} & \textbf{99.25(1.11)}  & 96.70(1.46) & 95.56(1.96) & \textbf{99.20(0.76)} & 96.88(1.80) & 100.00(0.01) \\
   \hline         
        \textbf{Haberman} & \textbf{76.71(1.86)} & 75.65(2.68) & 72.63(3.45) & 75.26(3.38) & 73.15(3.68) & 97.36(3.34) \\
\hline
	    \textbf{Cardiotocography (CTG)} & \textbf{84.62(1.08)} &  84.21(1.10) & \textbf{84.54(1.46)} & 83.06(1.23) & 84.04(2.02) & 93.08(1.46)  \\		
 \hline
       \textbf{Vertebral Column} & \textbf{86.89(2.46)} &  82.04(2.17) &  85.89(3.47) & 83.22(3.59) & 84.27(3.24) & 97.40(0.54)    \\		
\hline 
       \textbf{Steel Plate Faults} & \textbf{67.21(1.20)} &  66.05(1.98) &  67.02(1.98) & 66.57(1.06) & 67.22(1.64) & 88.72(1.89) \\	
\hline 
       \textbf{WDG V1} & \textbf{84.56(0.36)} &  83.17(0.76) &  84.36(0.56) & 84.04(0.37) & 84.23(0.53) & 97.82(0.54)  \\	
 \hline
       \textbf{Ecoli} & \textbf{77.25(3.52)} $\bullet$ &  69.35(2.68) & 72.22(3.65) & 70.32(3.65) & 67.80(4.60) & 91.54(1.55)   \\	
\hline	   
       \textbf{Glass} & \textbf{66.87(2.99)} $\bullet$ &  52.92(4.53) &  62.64(5.61) & 55.89(3.25) & 57.16(4.17) & 90.65(0.00)  \\	          
\hline		
       \textbf{ILPD} & \textbf{69.40(1.64)} &  67.53(2.83) &  67.20(2.35) & \textbf{69.38(4.28)} & 67.26(1.04) & 99.10(0.72)  \\	     
\hline
        \textbf{Adult} & \textbf{87.15(2.43)} $\bullet$ &  83.64(3.34) &  85.60(2.27) & 83.58(2.91) & 84.37(2.79) & 95.59(0.39)  \\	 
\hline
       \textbf{Weaning} & \textbf{79.67(3.78)} $\bullet$ &  74.86(4.78) &  76.31(4.06) & 74.47(3.68) & 76.89(3.15) & 92.10(0.92)  \\	
\hline
        \textbf{Laryngeal1} & \textbf{83.43(4.50)} &  80.18(5.51) &  81.32(3.82) & 79.81(3.88) & 80.75(4.93) & 98.86(0.98)  \\	
\hline
	    \textbf{Laryngeal3} & \textbf{72.65(2.17)} &  68.42(3.24) &  67.13(2.47) & 62.32(2.57) & 71.23(3.18) & 100(0.00)  \\	
\hline
       \textbf{Thyroid} & \textbf{96.78(0.87)} &  95.15(1.74) & 95.25(1.11) & 96.01(0.74) & 96.24(1.25) & 99.88(0.36)  \\	
\hline	
        \textbf{German credit} & \textbf{75.55(2.31)} &  71.16(2.39) &  74.76(2.73) & 72.96(1.25) & 73.60(2.69) & 99.12(0.70)  \\	
\hline
        \textbf{Heart} & \textbf{84.80(3.36)} &  80.26(3.58) &  82.50(4.60) & 81.61(5.01) & 82.05(3.72) & 95.90(1.02)  \\	
\hline        
        \textbf{Satimage} & \textbf{96.21(0.87)} &  94.52(0.96) & 95.23(0.87) & 95.43(0.92) & 95.31(0.92) & 98.69(0.87)  \\	
\hline
        \textbf{Phoneme} & \textbf{80.35(2.58)} $\bullet$ &  75.87(1.33) &  72.60(2.33) & 75.90(1.06) & 72.70(2.32) & 99.34(0.24) \\	
\hline		
		\textbf{Monk2} & \textbf{83.24(2.19)} &  79.25(3.78) & 79.18(2.57) & 80.27(2.76) & 80.55(3.59) & 98.98(1.19)  \\	
\hline		          
		\textbf{Mammographic} & 84.82(1.55) &  83.60(1.85) & \textbf{85.27(1.85)} & 83.07(3.03) & 84.23(2.14) & 99.59(0.15)  \\	
\hline		          
    	\textbf{MAGIC Gamma Telescope} & 84.35(3.27) &  80.27(3.50) &  81.24(2.22) & \textbf{87.35(1.45)} $\bullet$ & 85.25(3.25) & 95.35(0.68) \\	
    \hline
    \end{tabular}
} 
\end{table}

We can see in Table~\ref{table:Results} the proposed META-DES achieves results that are either superior or equivalent to the state-of-the-art DES techniques in 25 datasets (84\% of the datasets). In addition, the META-DES achieved the highest recognition performance for 18 datasets, which corresponds to 60\% of the datasets considered. Only for the Ecoli, Heart, Vehicle, Banana and Lithuanian datasets (16\% of the datasets) the recognition rates of the proposed META-DES framework presented is statistically inferior to the best result achieved by state-of-the-art DES techniques. 

\begin{figure}[H]
    \begin{center}
    	  \epsfig{file=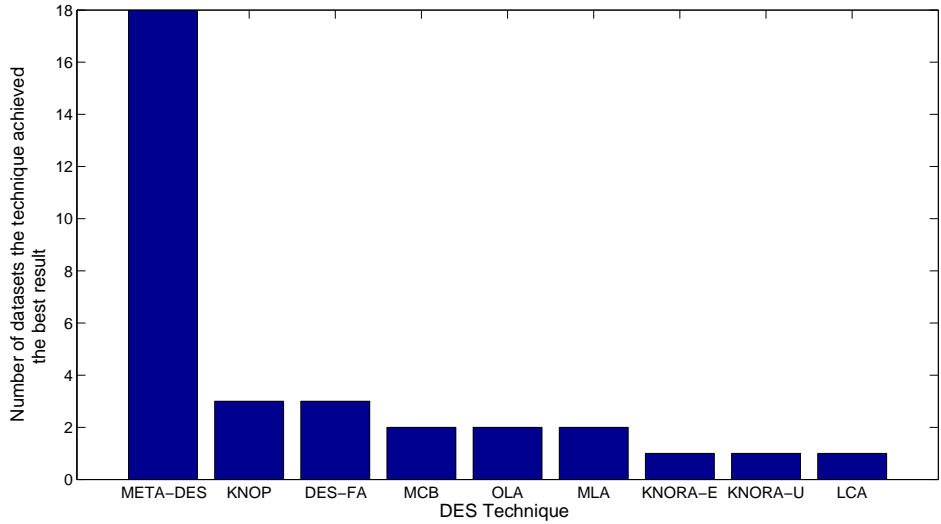, clip=, width=.80\textwidth}
    	  \end{center}
          \caption{Bar plot showing the number of datasets that each DES technique presented the highest recognition accuracy.}
    \label{fig:barresults}
\end{figure}

For the $12$ datasets where the proposed META-DES did not achieved the highest recognition rate (WDBC, Banana, Vehicle, Lithuanian, Cardiotocography, Vertebral column, Steel plate faults, Ecoli, Glass, ILPD, Laryngeal3 and Heart) we can see that each DES technique presented the best accuracy for different datasets (as shown in Figure~\ref{fig:barresults}). The KNOP achieves the best results for three datasets (Ecoli, Steel plate faults and Laryngeal3), the MCB for two datasets (Vehicle and Glass), the DES-FA for 3 datasets (Banana, Breast cancer and Cardiotocography) and so forth. This can be explained by the "no free lunch" theorem. There is no criterion to estimate the competence of base classifiers that dominates all other when compared with several classification problems. Since the proposed META-DES uses a combination of five different criteria as meta-features, even though one criterion might fail, the system can still achieve a good performance as other meta-features are also considered by the selection scheme. In this way, a more robust DES technique is achieved.

Moreover, another advantage of the proposed META-DES framework comes from the fact that several meta-feature vectors are generated for each training sample in the meta-training phase (Section 3.2.2). For instance, consider that $200$ training samples are available for the meta-training stage ($N = 200$), if the pool $C$ is composed of $100$ weak classifiers ($M = 100$), the meta-training dataset is the number of training samples $N$ $\times$ the number classifiers in the pool $M$, $N \times M = 20.000$. Hence, there is more data to train the meta-classifier $\lambda$ than for the generation of the pool of classifiers $C$ itself. Even though the classification problem may be ill-defined, due to the size of the training set, using the proposed framework we can overcome this limitation since the size of the meta-problem is up to 100 times bigger than the classification problem. So, our proposed framework has more data to estimate the level of competence of base classifiers than the other DES methods, where only the training or validation data is available. This fact can be observed by the results obtained for datasets with less than 500 samples for training, such as, Liver Disorders, Sonar, Weaning and Ionosphere where recognition accuracy of the META-DES is statistically superior for those small size problems.

When compared against static ensemble techniques Table~\ref{table:Results2}, the proposed META-DES achieves the highest recognition accuracy for 24 out of 30 datasets. This can be explained by the fact that the majority of datasets considered are ill-defined. Hence, the results found in this paper also support the claim made by Cavalin et al.~\cite{paulo2} that DES techniques outperform static methods for ill-defined problems.

We can thus answer the research question posed in this paper: Can the use of meta-features lead to a more robust dynamic selection technique? As the proposed system achieved better recognition rates in the majority of datasets the use of multiple properties from the classification environment as meta-features indeed leads to a more robust dynamic ensemble selection technique.

\section{Conclusion} 
\label{sec:conclusion}  

In this paper, we presented a novel DES technique in a meta-learning framework. The framework is based on two environments: the classification environment, in which the input features are mapped into a set of class labels, and the meta-classification environment, in which different properties from the classification environment, such as the classifier accuracy in the feature space or the consensus in the decision space, are extracted from the training data and encoded as meta-features. Five sets of meta-features are proposed. Each set corresponding to a different dynamic selection criterion. These meta-features are used to train a meta-classifier which can estimate whether a base classifier is competent enough to classify a given input sample. With the arrival of new test data, the meta-features are extracted using the test data as reference, and used as input to the meta-classifier. The meta-classifier decides whether the base classifier is competent enough to classify the test sample. 

Experiments were conducted using 30 classification datasets coming from five different data repositories (UCI, KEEL, STATLOG, LKC and ELENA) and compared against eight state-of-the-art dynamic selection techniques (each technique based on a single criterion to measure the level of competence of a base classifier), as well as five classical static combination methods. Experimental results show the proposed META-DES achieved the highest classification accuracy in the majority of datasets, which can be explained by the fact that the proposed META-DES framework is based on five different DES criteria. Even though one criterion might fail, the system can still achieve a good performance as other criteria are also considered in order to perform the ensemble selection. In this way, a more robust DES technique is achieved. 

In addition, we observed a significant improvement in performance for datasets with critical training size samples. This gain in accuracy can be explained by the fact that during the Meta-Training phase of the framework, each training sample generates several meta-feature vectors for the training of the meta-classifier. Hence, the proposed framework has more data to train the meta-classifier and consequently to estimate the level of competence of base classifiers than the current state-of-the-art DES methods, where only the training or validation data is available.

 
Future works on this topic will involve:

\begin{enumerate}
\item The definition of new sets of meta-features to better estimate the level of competence of the base classifiers.
\item The selection of meta-features based on optimization algorithms in order to improve the performance of the meta-classifier, and consequently, the accuracy of the DES system.
\item The evaluation of different training scenarios for the meta-classifier.
\end{enumerate}

\section*{Acknowledgment}
This work was supported by the Natural Sciences and Engineering Research Council of Canada (NSERC), the \'{E}cole de technologie sup\'{e}rieure (\'{E}TS Montr\'{e}al) and CNPq (Conselho Nacional de Desenvolvimento Cient\'{i}fico e Tecnol\'{o}gico).

\bibliographystyle{elsarticle-num}
\bibliography{report}

\end{document}